\documentclass[sigconf]{acmart}

\AtBeginDocument{%
  \providecommand\BibTeX{{%
    \normalfont B\kern-0.5em{\scshape i\kern-0.25em b}\kern-0.8em\TeX}}}
\usepackage{multirow}
\usepackage{xcolor}

\usepackage{natbib} 
\usepackage{microtype}
\usepackage{graphicx}
\usepackage{booktabs} %
\usepackage{tabularx}
\usepackage{multirow}
\usepackage{pifont}
\usepackage{tikz}
\usetikzlibrary{shapes,arrows,decorations.markings}
\usetikzlibrary{positioning}
\usepackage{pgfplots} 
\pgfplotsset{compat=newest} 
\usetikzlibrary{plotmarks}
\usetikzlibrary{arrows.meta}
\usepgfplotslibrary{patchplots}
\newlength\plotH
\newlength\plotW
\newlength\plotHthreecol
\usepackage{hyperref}
\usepackage{microtype}
\usepackage{graphicx}
\usepackage{booktabs} 
\usepackage{natbib}

\usepackage{amssymb}
\usepackage{amsthm}
\usepackage{lipsum}
\usepackage{xspace}
\usepackage{float}
\usepackage{multirow}
\usepackage{microtype}
\usepackage{booktabs}
\usepackage{enumitem}
\usepackage[utf8]{inputenc}
\usepackage{listings}
\usepackage{caption}
\usepackage{dsfont}
\usepackage{algpseudocode}
\usepackage{graphicx}
\usepackage{subfig}
\usepackage{amsthm}
\usepackage{afterpage}
\usepackage{xspace}
\usepackage{adjustbox}
\usepackage{ragged2e} 
\usepackage{booktabs, tabularx}
\usepackage{makecell}
\usepackage{nicematrix,tikz}
\usepackage{flushend}
\usepackage{hyperref}
\newcommand{\ours}{ECCW\xspace}
\newcommand{\Expert}{\text{FFN}}

\newcommand{\combine}{\mathcal{G}}  

\newcommand{\WI}{wi}
\newcommand{\WO}{wo}
\usepackage{amsmath}
\usepackage{amssymb}
\usepackage{mathtools}
\usepackage{amsthm}
\usepackage{xfrac}
\usepackage{float}
\usepackage{stfloats}

\acmSubmissionID{****}
\renewcommand\footnotetextcopyrightpermission[1]{}

\begin{document}
\fancyhead{}
\title{The Future of Combating Rumors? Retrieval, Discrimination, and Generation}

\author{Junhao Xu}
\authornote{corresponding author. Email:<xujunhao1105@gmail.com>}
\affiliation{%
  \institution{University of Malaya}
   \city{Kuala Lumpur}
   \country{Malaysia}
}

\author{Longdi Xian}
\affiliation{%
  \institution{The Chinese University of Hong Kong}
  \city{Hong Kong}
  \country{China}
}

\author{Zening Liu}
\affiliation{%
  \institution{University of Malaya}
   \city{Kuala Lumpur}
   \country{Malaysia}
}

\author{Mingliang Chen}
\affiliation{%
  \institution{Peking University}
   \city{Shen Zhen}
   \country{China}
}

\author{Qiuyang Yin}
\affiliation{%
  \institution{Guangzhou College of Technology and Business}
   \city{Guang Zhou}
   \country{China}
}
\author{Fenghua Song}
\affiliation{%
  \institution{Guangzhou College of Technology and Business}
   \city{Guang Zhou}
   \country{China}
}
\renewcommand{\shortauthors}{Xu}
\newcommand{\tabincell}[2]{\begin{tabular}{@{}#1@{}}#2\end{tabular}} 

\begin{abstract}
Artificial Intelligence Generated Content (AIGC) technology development has facilitated the creation of rumors with misinformation, impacting societal, economic, and political ecosystems, challenging democracy. Current rumor detection efforts fall short by merely labeling potentially misinformation (classification task), inadequately addressing the issue, and it is unrealistic to have authoritative institutions debunk every piece of information on social media. Our proposed comprehensive debunking process not only detects rumors but also provides explanatory generated content to refute the authenticity of the information. The Expert-Citizen Collective Wisdom (\ours) module we designed aensures high-precision assessment of the credibility of information and the retrieval module is responsible for retrieving relevant knowledge from a Real-time updated debunking database based on information keywords. By using prompt engineering techniques, we feed results and knowledge into a LLM (Large Language Model), achieving satisfactory discrimination and explanatory effects while eliminating the need for fine-tuning, saving computational costs, and contributing  to debunking efforts.
\end{abstract}



\maketitle

\section{Introduction}
\begin{figure}[ht]
    \centering
    \includegraphics[width=0.9\linewidth]{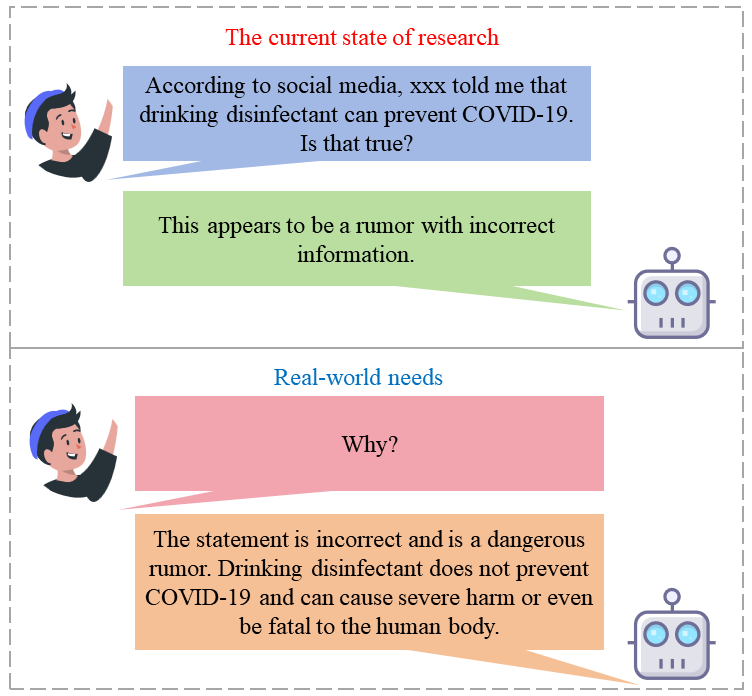}
    \caption{Example of rumor. Both in the academic and industrial sectors, only the first half of the work has been done so far. However, for users influenced by information, this is not sufficient to convince them. Therefore, the second half of the work is what reality demands.}
    \label{fig:rumor}
\end{figure}
Social media platforms such as Weibo\footnote{https://m.weibo.cn} and Twitter\footnote{ https://twitter.com} have become increasingly popular for users to share information, express their opinions, and keep up with current events \cite{pham2022bot2vec}. However, these platforms have also been exploited by some individuals to spread rumors. Rumors are defined as widely spread stories or statements that have not been verified or confirmed for their truthfulness \cite{yang2023rumor}. Given the sheer number of users on social media and the ease with which information can be accessed and shared, unverified reports and fabricated rumors can spread quickly, potentially causing widespread confusion and panic within society, and even having economic and political consequences \cite{bi2022microblog}. In order to address this issue, putting an end to rumors on social media is crucial to mitigate their potential negative impact.

In the past, the primary focus of rumor research centered around rumor classification \cite{dou2021user,yang2023rumor,zheng2023rumor} and rumor verification \cite{you2019attributed}. While these studies yielded promising results on their respective datasets, the models developed therein were predominantly designed as single-task systems, specializing in tasks such as classification and regression. However, these models often fell short of fully leveraging the potential of language, and they lacked the expressive and interactive capabilities highlighted in recent research 
 \cite{singhal2023large}. As a result, a notable disparity exists between the current capabilities of these models and the expectations placed on them within real-world rumor debunking workflows (See Figure \ref{fig:rumor}). A mature rumor debunking system requires diverse capabilities. First, it needs to accurately comprehend the complexity of human language \cite{cui2023intra,nguyen2020fang}, including semantics, logic, and emotions. Only such a system can make precise judgments when faced with various forms of rumors and identify any loopholes or contradictions \cite{lin2023zero}. Secondly, a successful rumor debunking system should integrate a vast amount of existing knowledge and information to enhance its ability to discern the authenticity of rumors \cite{cui2020deterrent}. This requires the system to gather data from multiple sources and carry out effective information integration and reasoning to form comprehensive and reliable rumor debunking results \cite{sun2023inconsistent}. Furthermore, the rumor debunking system should be interactive, allowing real-time communication and interaction with users \cite{singhal2023large}. In practical applications, the rumor debunking system must also consider the timeliness and reliability of information \cite{zhu2022generalizing}. As rumors can spread rapidly on social media, providing timely feedback and responses are crucial to prevent further dissemination of misinformation.

With the rise of large language models and the successful applications like ChatGPT\footnote{https://chat.openai.com/} , new possibilities have emerged for the development of automated and interactive rumor debunking systems. The significance of large language models in the field of natural language processing cannot be underestimated. Their powerful representation learning abilities enable them to grasp the complexities and semantic information of language from extensive corpora \cite{vaswani2017attention}. As a result, these models provide a robust foundation for the construction of efficient rumor debunking systems.The paper presents contributions concerning the integration of rumor debunking systems with large language models. Our contributions are outlined below:
\begin{itemize}
\item To the best of our knowledge, we are the first to develop a comprehensive rumor debunking system using a large language model (involving retrieval, discrimination, and guided generation). 
\item We proposed \ours and evaluated its discriminative performance against rumors from various domains. The results show that our model, by encoding textual information and feeding it into two parallel subnetworks, achieves the optimal discriminative effect.
\item We have transformed existing publicly debunked content into a searchable knowledge vector database. This database can be utilized for keyword-based retrieval of information related to relevant rumor queries, facilitating the provision of the reasoning and analytical knowledge required for practical debunking applications.
\item We integrate the discriminative results from \ours with keyword information retrieved from a vector database through the prompt engineering process. Utilizing  the Retrieval Augmented Generation, we have successfully produced persuasive information that debunks rumors.
\end{itemize}

\section{Related Work}
Before the advent of transformer, some rumor detection efforts primarily relied on statistical methods \cite{xu2021unified}. However, in the era of Large Language Models (LLMs), these methods have become almost ineffective, as LLMs like GPT can be exploited to generate rumors for sensationalism. The research indicated that, compared to rumors containing misinformation crafted by humans, those generated by LLMs are more challenging for both humans and detectors to identify \cite{ji2023survey}. This suggests that they may exhibit a more deceptive style and potentially cause greater harm \cite{anonymous2024can}.At the same time, the emergence of LLMs holds tremendous potential for reshaping the landscape in combating misinformation. Their extensive and diverse global knowledge, coupled with robust reasoning capabilities, presents a promising opportunity to counteract misinformation.
\citet{2021Fake} studied the use of large language models (LLMs) to explore the interaction between machine-generated and human-written real and fake news articles. They found that detectors trained specifically on human-written articles were indeed good at detecting machine-generated fake news, but not vice versa. \citet{pavlyshenko2023analysis} explored the possibility of using the Llama 2 large language model (LLM) based on PEFT/LoRA fine-tuning for fake news detection and information distortion analysis. \citet{chen2023can} investigated the understanding capabilities of multiple LLMs regarding both content and propagation across social media platforms. They designed four instruction-tuned strategies to enhance LLMs for both content and propagation-based misinformation detection, resulting in better detection performance.

The detection task is merely one component of the comprehensive debunking process; it is essential to include a fact-checking step to assist users in comprehending inaccuracies within the information. A deep-learning-based fact-checking URL recommender system was suggested with the aim of offering dependable information sources for inquiring users \cite{you2019attributed}. Nevertheless, a drawback of this approach is that the majority of users prefer content that is direct and explanatory, rather than having to open websites and manually search for relevant evidence. Therefore, utilizing the generative capability of LLMs to produce explanatory content for information is a reasonable direction. However, the hallucination issues of LLMs may result in the generated explanatory content containing a significant amount of misinformation \cite{chen2023combating}, turning the inquirer into a secondary victim of rumors. In order to reduce the problem of hallucinations caused by the LLM model, \citet{wei2022chain} employed a Chain of Thought (CoT) prompt to induce more faithful reasoning. \citet{jiang2023structgpt} used knowledge graphs, datasheets, and structured databases as external knowledge bases to improve LLM reasoning. \citet{tian2023graph} pointed out that utilizing Knowledge Graphs (KGs) to enhance language modeling through joint training and custom model architectures incurs significant costs. Retrieval-augmented generation, however, could be a better solution. They proposed Graph Neural Prompting, which demonstrated advantages in commonsense and reasoning tasks across various scales and configurations of LLMs.

\section{Method}
\begin{figure*}[h]
    \centering
    \includegraphics[width=\linewidth]{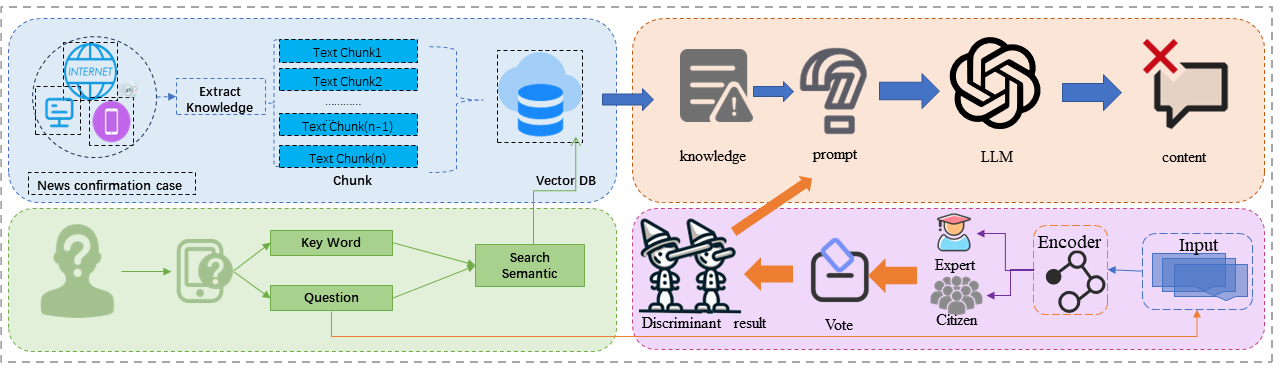}
    \caption{The comprehensive framework for debunking processes.In the bottom-left corner, it represents the user's inquiries about the authenticity of the rumors they need to verify. In the top-left corner lies the process of constructing a database of debunking knowledge vectors. The bottom-right corner depicts the simplified workflow of our rumor discrimination network. The top-right corner illustrates how to obtain interpretable debunking content by integrating the debunking results with the retrieved knowledge through RAG (Retrieval-Augmented Generation).}
    \label{fig:1}
\end{figure*}
In this section, we will introduce the debunking process we proposed and provide a detailed explanation of its components and design, including discriminative networks, retrieval, and generation methods. Figure \ref{fig:1} illustrates the comprehensive framework of our approach. Our discriminative networks, denoted as \ours, are employed for the classification and discrimination of rumor-related information. These networks incorporate Semantic Encoding to encode the raw text initially. Subsequently, the text undergoes processing through two subsidiary networks. 1) Domain Expert Discrimination: This is utilized for semantic decomposition tasks. By employing gating mechanisms, it directs semantics to domain-specific experts, thus enhancing accuracy and saving computational resources. 2) Citizen Perceptual: This is employed to simulate the diversity in individuals' comprehension of information, akin to introducing semantic noise. These semantics then traverse through a fully connected graph network to mimic interpersonal communication. The information captured by the graph network is concatenated with the original information to simulate the process of individuals reconsidering their viewpoints after discussions with others in real-life scenarios. Finally, the classification results from the two subsidiary networks are integrated using a voting mechanism named Collective Wisdom Decision to achieve classification effectiveness. The Retrieval module encompasses the construction of a vector database. This is essential for achieving precise and real-time debunking effects while minimizing the influence of large language model hallucinations. It is impractical to rely on fine-tuning large language models for debunking explanations due to the considerable time investment involved. However, maintaining and updating a debunking database in real-time is a viable solution. When users pose queries, relevant knowledge can be swiftly retrieved from the vector database through keyword matching, enabling rapid verification of information authenticity. Generation methods involve amalgamating classification results with retrieved knowledge using prompt techniques. This ensures not only the discernment of rumor veracity but also the provision of comprehensive explanations, thereby achieving a comprehensive debunking effect throughout the entire process.

\subsection{Semantic Encoding}
We use  a pre-training model to generate text embeddings, which can better capture semantic information in textual data and provide more powerful feature representation for subsequent classification tasks. LLM has a deeper understanding of linguistics, resulting in a higher expressive capability in text representation. Here, we set the sequential text as $T_i$, and the final sentence  vector is $S_i$. The  embedding of the text is represented by the following formula:
\begin{equation}
  S_i = Transformer(T_i),
\end{equation}
$\boldsymbol{S}_i\in\mathbb{R}^d$ is the representation of the textual information, where $d$ is the embedding size. In particular, we use Chinese llama \cite{cui2023efficient}, which, based on the original vocabulary of LLaMA, adds 20,000 Chinese tokens to help prevent Chinese text from being segmented into too many fragments, thereby improving encoding efficiency. We load a pre-trained Chinese LLaMA tokenizer. Each text is processed, and the tokenizer is used to convert the text into a format that the model can handle. Subsequently, we get the final embedding vector through the pre-trained model. This embedding vector can be used for subsequent machine learning tasks, such as classification, clustering, or other NLP-related applications. 

\subsection{Domain Expert Discrimination}
\begin{figure}[ht]
    \centering
    \includegraphics[width=0.98\linewidth]{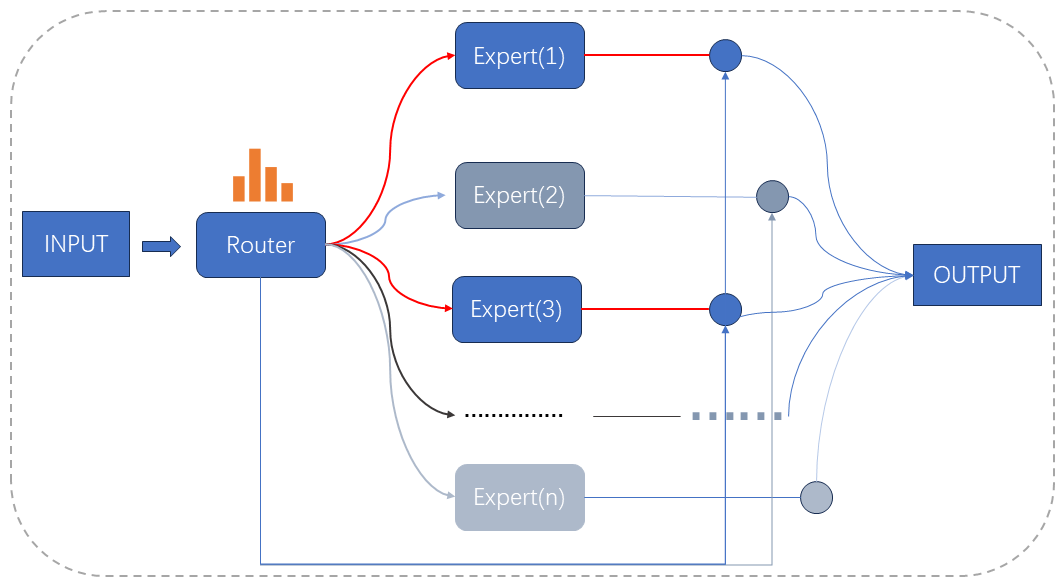}
    \caption{Expert Discrimination. The input will be routed through a router to allocate domain experts, and finally, the selected expert information will be summarized.}
    \label{fig:2}
\end{figure}
Due to the typically cross-domain nature of rumors, such as the convergence of financial and social science domains, it is imperative to enlist specialized domain experts for evaluating the professional credibility of pertinent domain information. To emulate this process, we introduce a hybrid expert network designed to channel semantic information into different domains, each overseen by respective experts (see Figure \ref{fig:2}). Specifically, a Router network is responsible for analyzing the domain category to which a given message belongs and submitting it for evaluation to domain-specific experts. Each expert, having gained a distinct understanding within their respective domains, is then subject to semantic assessments by an expert ensemble network responsible for consolidating these evaluations. The network defined as follows:
\begin{equation}
    \combine_{i,E}= Router(S_i),
\end{equation}
where $S_i$ is the $i$th text vector that has been encoded semantically. $\boldsymbol{\combine_{i,E}}\in\mathbb{R}^E$ is computed through a Router network, where each non-negative entry in $\combine_{i,E}$ corresponds to an expert. This vector directs the feature to a select few experts $E$, with the non-zero entries indicating the extent to which an expert contributes to the final network output.The specific computation is shown in the following equation:
\begin{equation}
    \\Router(x)=Softmax(x,W),
\end{equation}
where $\boldsymbol{W}\in\mathbb{R}^{d\times E}$ represents the weight coefficients of the routing network, where $d$ is the text dimension and $E$ is the number of experts configured.
\begin{equation}
    \Expert_e(S_i)= \WO_e \cdot \text{ReLU}(\WI_e \cdot S_i),    
\end{equation}
\begin{equation}
    H_i = \sum^E_{e=1} \combine_{i,e} \cdot \Expert_e(S_i).
\end{equation}
The input ${S}_i$ is processed, with $\WI_e$ and $\WO_e$ serving as the input and output projection matrices for the feed-forward layer (an expert), where $e$ denotes the index corresponding to the expert. The hybrid expert projection decision layer output, $H_i$, is then calculated as the weighted average of the outputs from the chosen experts.

\subsection{Citizen Perceptual}
\begin{figure}[ht]
    \centering
    \includegraphics[width=\linewidth]{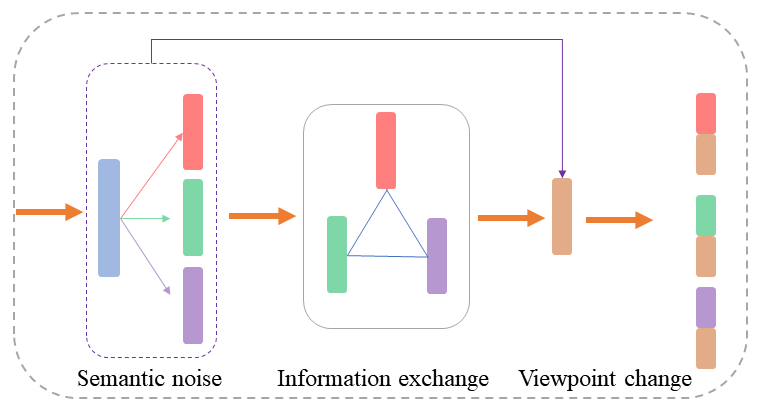}
    \caption{Citizen Perceptual. The input will first pass through semantic noise simulating different people's perspectives, then gain insights from information exchange with others' viewpoints, and finally integrate with one's initial perspective to form a revised viewpoint.}
    \label{fig:3}
\end{figure}
When it comes to discerning rumors, relying solely on expert opinions is insufficient. Even experts can make mistakes. Therefore, harnessing the power of the general public is necessary. We have devised a civic perception layer to simulate the diverse perceptions of individual citizens towards the given information (see Figure \ref{fig:3} ), as defined below:
\begin{equation}
    P_i^j=  {\phi}(S_i),
\end{equation}
\begin{equation}
    M_i^j=  {\varphi}(P_i^j),
\end{equation}
where $\phi(\cdot)$ is the citizen diversity resolver. Specifically, we configure n MLPs to map $S_i$, generating n citizen understanding vectors with subtle differences, and $j=1,2,...,n$. $\varphi(\cdot)$ is a quadratic encoding layer. We employ Bidirectional Long Short-Term Memory Network (BiLSTM), which can be substituted with other networks. The purpose of doing so is to simulate the diversity in understanding of the same information among different individuals by introducing information noise. Then, to enhance the robustness of the model, we aggregate these diversified understandings of the same information through an information discussion network, defined as follows:
\begin{equation}\label{eq:1}
    N_i^j =  {Att(M_i^j,{A})},
\end{equation}
where ${A}$ corresponds to the adjacency matrix for the set of distinct semantic vectors $M_i$, simulating different citizens. Heuristically adjustable, the adjacency matrix can be tailored according to a specified strategy. In our methodology, we establish full connectivity, thereby constituting a globally interconnected graph that simulates the network of interactions and discussions among citizens. Implementing our global graph network involves incorporating an  attention mechanism $Att(\cdot)$, as outlined below:
\begin{equation}
    {Att(M_i^j,{A})}   =   \text{softmax}\left({M}_Q{M}_K^T\right){M}_V.
\end{equation}
In the softmax mechanism, the terms $Q$, $K$, and $V$ correspond to the query, key, and value matrices, respectively. Their pivotal role within this mechanism is to facilitate the computation of attention scores.
$Q$ signifies information from the current node, serving as the basis for deriving attention scores that govern interactions with other nodes. On the other hand, $K$ encapsulates details about neighboring nodes, contributing to the determination of the relevance or importance of each neighbor in relation to the current node. Meanwhile, $V$ embodies the actual information associated with each neighboring node.

The softmax operation applied to the matrix product ${M}_Q{M}_K^T$ essentially calculates attention scores, determining the relative importance of different neighbors based on the compatibility of their features. Subsequently, these attention scores are utilized to weigh the corresponding values ($V$), yielding an aggregated representation that accentuates more pertinent information from the neighbors within the broader context of the graph.

Subsequently, we concatenate the citizens' initial cognitions with those refined through network processing, culminating in the ultimate decision cognition, defined as follows:
\begin{equation}\label{eq:2}
    {C}_i^j = \text{concat}\left({P}_i^j, {N}_i^j\right).
\end{equation}
This concatenation approach combines the initial cognition with information refined through network processing, simulating the integration of individual understanding, obtained through autonomous cognition, and the collective understanding resulting from discussions with others, thereby reducing information loss and comprehensively considering information from various sources. It helps incorporate a more comprehensive and multi-faceted range of information into the final decision-making cognition.

\subsection{Collective Wisdom Decision}
After obtaining distinct decision cognitions from both experts and citizens, it is necessary to compute their final choices. Here, we employ softmax for this purpose, defined as follows:
\begin{equation}
    {E}_i = \text{softmax}(H_i),
\end{equation}

\begin{equation}
    {D}_i^j = \text{softmax}(C_i^j),
\end{equation}
where $H_i$ and $C_i^j$ are derived from equations \eqref{eq:1} and \eqref{eq:2} respectively, while ${E}_i$ and ${D}_i^j$ represent their decision outcomes. To further refine the decision-making process, we introduce a voting mechanism that aggregates the decision outcomes from both experts and citizens. This involves setting up a voting box, where each individual's decision, represented by ${E}_i$ and ${D}_i^j$ for experts and citizens respectively, contributes to the final decision through a democratic voting approach:
\begin{equation}
    {F_i} = {softmax}\left(W_e \cdot E_i + W_c \cdot D_i^j\right),
\end{equation}
where $W_e$ represents the weight parameter matrix for experts, indicating their influence on the final decision. $W_c$ represents the weight parameter matrix for citizens, indicating their influence on the final decision. The introduction of this voting mechanism facilitates the achievement of collective wisdom by blending the perspectives of both experts and citizens into the final decision.

\subsection{Retrieval-Augmented Generation}
Due to the timeliness of rumors, conducting real-time fine-tuning on LLMs based on new corpora would consume significant computational resources, which is impractical. Therefore, we consider adopting the RAG approach, where by updating the vector database in real-time, we can reduce the impact of hallucination while generating more accurate and persuasive debunking content, without sacrificing the generalization ability of LLMs in the domain.

\textbf{Vector database}. The data source for building the knowledge base is the debunking content contained in the rumor dataset used in this experiment, ensuring the reliability of the knowledge source itself. We chunk all knowledge texts and then embed them, storing them in vector form for easy retrieval queries.In detail, we set the chunk size to 100, and utilize OpenAI's official embedding method for ease of compatibility with future GPT-4 models.

\textbf{Knowledge retrieval}.
The information queries raised by users, which require confirmation, will also utilize the same embedding method used in constructing the vector database.The semantic search \cite{he2023efficient} involves comparing the question embedding with the database of text chunk embeddings:
\begin{equation}
S(Q, T_i) = \frac{Q \cdot T_i}{\|Q\|\|T_i\|},
\end{equation}
where $Q$ represents the question embedding, and $T_i$ signifies the embedding of the $i^{\text{th}}$ text chunk. Subsequently, the system ranks the text chunks according to their similarity scores. From this ranked list, the top N text chunks with the highest similarity scores are selected. These chunks are deemed the most relevant to the user's query.

\textbf{Content generation}. 
After obtaining the discriminative result of the information (the truthfulness or uncertainty of the rumor), as well as the relevant knowledge retrieved from the vector database, we proceed to the final Retrieval-Augmented Generation through Prompt Engineering:
\begin{equation}
Prompt = \psi(F,K,Q),
\end{equation}
where $F$ denotes the discriminative outcome of \textit{\ours} regarding the rumor, $K$ signifies the retrieved relevant knowledge, and $Q$ represents the original information of the queried rumor. $\psi(\cdot)$ represents specific prompting techniques, such as CoT (Chain-of-Thought)  and others.

After undergoing thorough prompt engineering, utilizing the final fused content as input for Large Language Models (LLMs) allows for obtaining debunking information, thereby effectively combating the spread of rumors containing false information.

\section{Experiments}
\subsection{Dataset}
The dataset used in this study is CHEF \cite{hu-etal-2022-chef}, the first Chinese Evidence-based Fact-checking dataset consisting of 10,000 real-world claims. The dataset spans across multiple domains, ranging from politics to public health, and provides annotated evidence retrieved from the internet.

\begin{table}[ht]
    \centering
    \scriptsize
    \caption{Comparisons of fact-checking datasets. Source means where the evidence are collected from, such as fact-checking websites (FC). Retrieved denotes if the evidence is given or retrieved from the source. Annotated means whether the evidence is manually annotated.}
    \begin{NiceTabular}{c|c|c|c|c|c}
    \toprule
    Dataset & Domain & \#Claims &  Source & Retrieved & Annotated \\
    \midrule
    \multirow{1}{*}{Liar} & Multiple &12836 &FC & \textcolor{red}{\textbf{X}} &\textcolor{red}{\textbf{X}} \\
    \midrule
    \multirow{1}{*}{PolitiFact} & Politics &106 &FC &\textcolor{red}{\textbf{X}}  & \textcolor{red}{\textbf{X}}  \\
    \midrule
    \multirow{1}{*}{XFact} & Multiple &31189 &Internet & \textbf{\textcolor{teal}{\checkmark}} & \textcolor{red}{\textbf{X}} \\
    \midrule
    \multirow{1}{*}{CHEF} & Multiple & 10000& Internet& \textcolor{teal}{\checkmark} & \textbf{\textcolor{teal}{\checkmark}}  \\
    \bottomrule
    \end{NiceTabular}
    \label{tab:1}
\end{table}
The reason for selecting only CHEF as our dataset is illustrated in Table \ref{tab:1}. Firstly, its multiple domains ensure information diversity. Secondly, its source being the internet meets the characteristic of widespread rumor dissemination. Most importantly, CHEF's claims have already been retrieved and annotated, ensuring data reliability. Other publicly available datasets in the realm of rumors or fake information fail to meet such stringent criteria.

\subsection{Baselines}
During the discrimination phase, we used multiple pre-trained language models directly connected to a classifier as baselines (Unless weights pretrained on Chinese datasets are unavailable, we will prioritize using weights pretrained on Chinese datasets.), and focused on the performance on rumor data sets in six different fields, using accuracy, precision, recall and F1 scores as evaluation indicators. These discrimination baseline models are:
{BERT} \cite{devlin2018bert},
{ALBERT} \cite{lan2019albert},
{BART} \cite{lewis2019bart},
{XLNet} \cite{yang2019xlnet}, 
{RoBERTa} \cite{liu2019roberta}, 
{DeBERTa} \cite{he2020deberta} and 
{LLaMA2} \cite{touvron2023llama}.

In the generation phase, we utilized the GPT-4 model \cite{achiam2023gpt}, and concurrently, we compared various prompting techniques, including zero-shot \cite{wei2021finetuned}, few-shot \cite{workrethinking}, and CoT \cite{wei2022chain}, both with and without RAG \cite{lewis2020retrieval}.

\subsection{Implementation Details}
Our experiment runs on a T4 GPU, with our \ours model having 768 hidden layers, a batch size of 1024, a learning rate of 0.0001, and Adam used as the optimizer. Additionally, the training and testing datasets are divided in an 8:2 ratio.In knowledge retrieval, we configure to retrieve only the most relevant chunk. The model used for generation is GPT-4, with the temperature set to 0.

\subsection{Performance Comparison}
\begin{table*}[ht]
\centering
\caption{Performance comparison of different models and domains. Domain represents the subdivision of the dataset into subset datasets for each specific field, Total denotes the entire dataset, Acc stands for Accuracy, Pre represents Precision, Rec stands for Recall, and F1 signifies the F1 score.}
\label{tab:2}
\small 
\begin{NiceTabular}{c|c|c|c|c|c|c|c|c|c}
\toprule
Domain & Metric & {BERT \cite{devlin2018bert}} & {ALBERT \cite{lan2019albert}} &{BART \cite{lewis2019bart}} & {XLNet \cite{yang2019xlnet}} & {RoBERTa \cite{liu2019roberta}} & {DeBERTa \cite{he2020deberta}} & {LLaMA2 \cite{touvron2023llama}}  & \textbf{\ours}\\
\midrule
 \multirow{4}{*}{Political} & Acc &81.57 &78.80 &56.22 &81.11 &65.90 &53.92 & 85.25&89.86\\
 & Pre &77.43 &53.67 &33.17 &69.45 &51.45 &40.94 &85.24 &89.78\\
 & Rec &61.71 &56.39 &33.60 & 63.19& 42.70&44.81 & 85.25&89.86\\
 & F1 &62.41 &54.65 & 26.29& 64.02 &40.49 &37.09 & 85.02&89.66\\
 \midrule
 \multirow{4}{*}{Culture} & Acc &75.16 & 72.64&66.98 &72.01 &66.98 &66.98 &85.53 &92.14\\
 & Pre &71.36 &72.40 &22.33 &71.16 &22.33 &22.33 &86.41 &92.40\\
 & Rec &69.11 &54.81 &33.33 &60.74 &33.33 &33.33 & 85.53&92.14\\
 & F1 &69.33 &58.50 &26.74 &64.52 &26.74 &26.74 & 85.27&92.12\\
 \midrule
 \multirow{4}{*}{Public health} & Acc &72.71  &67.51  & 52.88&67.51 &63.99 &60.34 & 81.58&84.11\\
 & Pre &63.70 &  43.81&34.58 &54.57 &39.92 & 20.11&82.08&83.13 \\
 & Rec & 54.57&46.81  &43.42 &47.46 &40.55 &33.33 &81.58 &84.11\\
 & F1 &50.84 &44.18 &36.85 &45.35 &37.40  &25.09 &77.81 &83.31\\
 \midrule
 \multirow{4}{*}{Society} & Acc &73.04 & 75.77&57.51 &67.58 &63.48 & 57.68 & 87.71&89.76\\
 & Pre &64.69 &69.72 &31.76 &59.84 &40.06 &52.46 & 86.74 &89.32\\
 & Rec &66.02 &63.60 &33.89 &62.74 &41.97 &34.08 &87.71& 89.76\\
 & F1 &63.98 &65.36 &25.48 &59.87 &39.04 &25.77 & 85.84&89.22\\
\midrule
 \multirow{4}{*}{Science} & Acc & 67.37&66.32 & 58.95&66.32 & 66.32&66.32 &85.26 &90.53\\
 & Pre &55.67 &22.11 & 30.91 &55.56 &22.11 & 22.11& 85.55&90.48\\
 & Rec &35.71 &33.33 &36.24 &34.66 & 33.33 &33.33  & 85.26&90.53\\
 & F1 &35.71 &26.58 & 33.36&30.00 &26.58 &26.58  &85.12 &90.30\\
\midrule
 \multirow{4}{*}{Life} & Acc & 100.0& 100.0&100.0 &50.00 &50.00 & 100.0&100.0 &100.0\\
 & Pre &100.0 &100.0 & 100.0&25.00 &25.00 &100.0 & 100.0&100.0\\
 & Rec &100.0 &100.0 &100.0 &50.00 &50.00 &100.0 &100.0 &100.0\\
 & F1 &100.0 &100.0 & 100.0& 33.33&33.33 &100.0 &100.0 &100.0\\
 \midrule
\multirow{4}{*}{Total} & Acc & 74.15 & 71.25 & 56.70 & 72.55 & 63.50 & 55.55 &85.05 &89.30\\
 & Pre & 65.62 & 64.46 & 39.57 & 64.55 & 62.59 & 40.82 &84.63 &89.11\\
 & Rec & 62.73 & 59.63 & 40.71 & 61.88 & 48.63 & 39.29 & 85.05&89.30\\
 & F1 & 63.40 & 60.48 & 35.98 & 62.57 & 47.49 & 33.49 &83.89& 89.19\\
\bottomrule
\end{NiceTabular}
\end{table*}
In this section, we analyze and compare the overall performance of \ours against selected baselines, as illustrated in Table \ref{tab:2}. For this experiment, we chose six distinct domains: politics, culture, public health, society, science, and lifestyle. A rapid overview of the results indicates that across all domains, the accuracy scores for \ours range from 84.11\% to 100.0\%, while the F-scores vary from 83.31\% to 100\%, outperforming other models. Additionally, when utilizing data from all domains, \ours achieves a 4.25\% higher accuracy and a 5.3\% higher F-score compared to the best-performing baseline model. This demonstrates the effectiveness of \ours in discerning the authenticity of rumors across multiple domains. Additionally, we observed that in the comparison of performance across domains, \ours achieved the highest performance in the Life domain (which had the smallest sample size relative to other domains). However, some models completely failed in this aspect, further illustrating how our collective wisdom approach can enhance effectiveness when dealing with limited-sample data. \ours exhibited the lowest discrimination accuracy for rumors in the Public Health domain, indicating that information in this domain is more prone to manipulation to deceive both the public and models. This further underscores the importance of integrating knowledge bases when generating explanatory content for this domain.

\begin{table*}[ht]
\centering
\caption{Performance comparison of different prompt engineering. RAG and Non-RAG represent whether to utilize the knowledge base. The highlighted red indicates the combination of the large language model with discrimination results and its own discernment to derive the classification outcome. The blue sections highlight some key conclusions derived from the large language model.}
\label{tab:3}
\small
\fontsize{6}{5}\selectfont
\resizebox{\linewidth}{!}{
\begin{NiceTabular}{c|m{1cm}<{\centering}|p{14cm}}
\hline
\toprule
Method  &Prompt & Result\\
 \midrule
  \multirow{26}{*}{None-RAG} & \multirow{4}{*}{Zero-shot} & \colorbox[RGB]{255,200,200}{\shortstack{This message is likely a rumor.}} 
  Firstly, \colorbox[RGB]{173,216,230}{\shortstack{donating one bag of blood does not harm a life}}. \colorbox[RGB]{173,216,230}{Blood donation is a highly safe process}, and the health of donors is thoroughly checked before donation. \colorbox[RGB]{173,216,230}{\shortstack{Only individuals meeting specific criteria are eligible to donate}}. Furthermore, the blood donation process utilizes disposable, sterile equipment, ensuring no harm to the donor.
  \\
  \\
&\multirow{6}{*}{Few-shot} & \colorbox[RGB]{255,200,200}{This message is rumor.} Relevant information indicates that blood donation organizations rigorously test and screen donated blood for infectious diseases such as HIV, hepatitis B, and hepatitis C. Only blood meeting strict standards in test results is used for clinical transfusions. Additionally, \colorbox[RGB]{173,216,230}{donors undergo health assessments before donation, excluding those with any risk factors that could compromise blood safety}. Therefore, \colorbox[RGB]{173,216,230}{accepting blood donation does not increase the risk of HIV infection}. \\
\\
&\multirow{12}{*}{CoT} & 
\colorbox[RGB]{255,200,200}{Conclusion: This message is likely a rumor.}
Argument:
Advanced modern medical technology includes rigorous screening before blood donation, including HIV testing. Even in rare cases of false negatives, blood undergoes secondary testing during storage to ensure safety.
\colorbox[RGB]{173,216,230}{Blood donation organizations typically assess donors' health, including their lifestyle and behavior, to evaluate the potential risk of HIV infection.} This is an additional protective measure to ensure the safety of the blood supply.
The statement "donating one bag of blood harms one life" is overly absolute. In reality, \colorbox[RGB]{173,216,230}{blood donation is a highly beneficial act, aiding those in need due to illness or injury}.
If accepting blood donation were genuinely hazardous, hospitals and medical institutions would not recommend and perform blood transfusions. However, in fact, blood transfusions are a crucial life-saving measure in many situations.
Lastly, the dissemination of such \colorbox[RGB]{173,216,230}{information could negatively impact public blood donation behavior}, \colorbox[RGB]{173,216,230}{affecting blood supply and harming those genuinely in need}. Hence, caution should be exercised in spreading misinformation. \\

\midrule
  \multirow{45}{*}{RAG}  & \multirow{8}{*}{Zero-shot} & \colorbox[RGB]{255,200,200}{This message might be a rumor.} Firstly, \colorbox[RGB]{173,216,230}{health departments}, including the Ministry of Health, adhere to scientific and medical principles when \colorbox[RGB]{173,216,230}{implementing blood donation policies to ensure safety}. Secondly, modern medicine has advanced blood testing technologies, effectively screening various viruses, including HIV. While \colorbox[RGB]{173,216,230}{no test guarantees 100\% accuracy}, nucleic acid testing is highly accurate, significantly reducing infection risks. Lastly, \colorbox[RGB]{173,216,230}{being a male who engages in same-sex activities does not equate to having HIV}. \colorbox[RGB]{173,216,230}{They have the right to participate in blood donation, provided their blood passes screening}. Therefore,this message may stem from \colorbox[RGB]{173,216,230}{misunderstandings and biases} against HIV and same-sex activity, rather than scientific facts. \\
  \\
 & \multirow{8}{*}{Few-shot} & \colorbox[RGB]{255,200,200}{This message may be a rumor.} Firstly, \colorbox[RGB]{173,216,230}{\shortstack{health departments}}, including the Ministry of Health, strictly adhere to scientific and medical principles when implementing blood donation policies to ensure safety. Secondly, modern medical technologies, such as \colorbox[RGB]{173,216,230}{\shortstack{nucleic acid testing}}, \colorbox[RGB]{173,216,230}{\shortstack{significantly reduce the detection errors of viruses in the blood}}. Although \colorbox[RGB]{173,216,230}{no testing method can guarantee 100\% accuracy}, rigorous testing and screening processes can minimize the risk of viral infection. Additionally, the claim "donating one bag of blood harms one life" is \colorbox[RGB]{173,216,230}{\shortstack{exaggerated and does not reflect the true nature of blood donation}}, which is \colorbox[RGB]{173,216,230}{a charitable act benefiting many patients in need of blood}. Therefore, denying the value of blood donation based on an extremely low risk is unwarranted. Finally, \colorbox[RGB]{173,216,230}{\shortstack{regarding the issue of blood donation by individuals with same-sex activities}}, as long as their blood test results are healthy, there is \colorbox[RGB]{173,216,230}{\shortstack{no reason}} to reject their donation. \colorbox[RGB]{173,216,230}{\shortstack{Sexual orientation does not determine one's health}}, and this is a fundamental human rights issue. \\
\\
& \multirow{10}{*}{CoT} & \colorbox[RGB]{255,200,200}{This message is likely a rumor based on assumptions and misconceptions.} Firstly, it assumes that \colorbox[RGB]{173,216,230}{blood from individuals with same-sex} activities carries a higher risk of HIV infection, which is an oversimplified and discriminatory viewpoint. Secondly, it \colorbox[RGB]{173,216,230}{misunderstands the accuracy of nucleic acid testing}. \colorbox[RGB]{173,216,230}{\shortstack{While no testing method is 100\% accurate}}, nucleic acid testing is one of the most precise methods for HIV detection.
Moreover, the message overlooks the blood testing and screening process. Before blood donation, all donors undergo comprehensive health checks and blood tests, including HIV testing. If the results are \colorbox[RGB]{173,216,230}{\shortstack{positive, the donated blood will not be used for transfusion}}.
Lastly, \colorbox[RGB]{173,216,230}{\shortstack{the message neglects the significance of blood supply}}. Blood donation is an act of saving lives, not causing harm. If certain individuals are deterred from donating blood due to fear and misunderstanding, it could lead to a shortage of blood supply, impacting patients in need of transfusions.
Therefore, \colorbox[RGB]{173,216,230}{\shortstack{this message appears to be a rumor rooted in fear and misunderstanding}} rather than factual information based on science. \\
\bottomrule
\end{NiceTabular}
}
\end{table*}

Table \ref{tab:3} presents and compares the performance of the GPT-4 model in Zero-shot, Few-shot, and CoT (Chain-of-Thought) settings for distinguishing rumors, and provides a detailed analysis of the results. Firstly, it is observed that the overall performance of the GPT-4 model is relatively poor in the Zero-shot setting. However, significant improvements in performance are observed in the Few-shot and CoT settings for the GPT-4 model. In the Few-shot setting, the model is able to learn from a small number of examples, thereby enhancing its ability to distinguish between true and false rumors. In the CoT setting, the model incorporates the coherence of information, understands the underlying logic of rumors, and makes more accurate judgments. This implies that the model can generate more precise textual outputs by integrating multiple relevant perspectives and arguments when dealing with true and false rumors.
Additionally, a comparison is made between the None-RAG (no knowledge) and RAG (with knowledge) settings. In the absence of a knowledge, the output of the GPT-4 model may be limited by its own training data and the capabilities of the language model. However, when employing the RAG setting, the model can leverage external knowledge to assist in judging true and false rumors. This enables the model to more accurately reference and apply domain-specific knowledge in its textual outputs, thereby enhancing precision.

\subsection{Ablation study}
\begin{table}[ht]
    \centering
    \scriptsize
    \caption{Performance comparison of ablation experiments.}
    \resizebox{\linewidth}{!}{
    \begin{NiceTabular}{c|c|c|c|c}
    \toprule
    Method & Acc & Pre & Rec & F1 \\
    \midrule
    \multirow{1}{*}{w/o Citizen \& Collective Wisdom} & 87.80 &84.62 &81.30 & 82.54 \\
    \midrule
    \multirow{1}{*}{w/o  Expert \& Collective Wisdom} & 87.02 &84.66 &79.88 & 81.47 \\
    \midrule
    \multirow{1}{*}{w/o Weighted voting} & 88.16 &85.31 &81.35 & 82.68 \\
    \midrule
    \multirow{1}{*}{full} & 89.30 & 89.11& 89.30&89.19  \\
    \bottomrule
    \end{NiceTabular}
    \label{tab:4}
}
\end{table}
Table \ref{tab:4} presents the outcomes of ablation experiments designed to assess the impact of removing specific components on the model's overall performance. We conduct a series of ablation studies to evaluate the effectiveness of the individual modules of the differents model. The specific ablation experiments
include the following aspects:\\
\textbf{w/o Citizen \& Collective Wisdom:}: Removing the Citizen and Collective Wisdom modules.\\
\textbf{w/o Expert \& Collective Wisdom:}: Removing the Expert and Collective Wisdom modules.\\
\textbf{w/o Weighted voting :}: Removing the Weighted Voting module.\\
Excluding both Citizen and Collective Wisdom from the scenario notably reduces the model's Accuracy, Precision, and other metrics, underscoring the substantial contributions of both Citizen and Collective Wisdom to the model's predictive capabilities. Likewise, omitting Expert and Collective Wisdom leads to a decrease in each metric, highlighting the critical role of Expert and Collective Wisdom in improving overall performance. The absence of Weighted Voting also leads to a decrease in various metrics. While Weighted Voting plays a role in performance, its impact appears to be less pronounced compared to other components.
Conversely, the full model, incorporating all components, achieves the highest overall performance with an Accuracy of 89.30\%, Precision of 89.11\%, Recall of 89.30\%, and F1 score of 89.19\%. This underscores the synergistic effects and complementarity of Citizen Perceptual, Domain Expert Discrimination, Collective Wisdom, and Weighted Voting in optimizing the model's predictive accuracy.

\subsection{Parameter Sensitivity}
\begin{figure}[ht]
    \centering
    \includegraphics[width=0.9\linewidth]{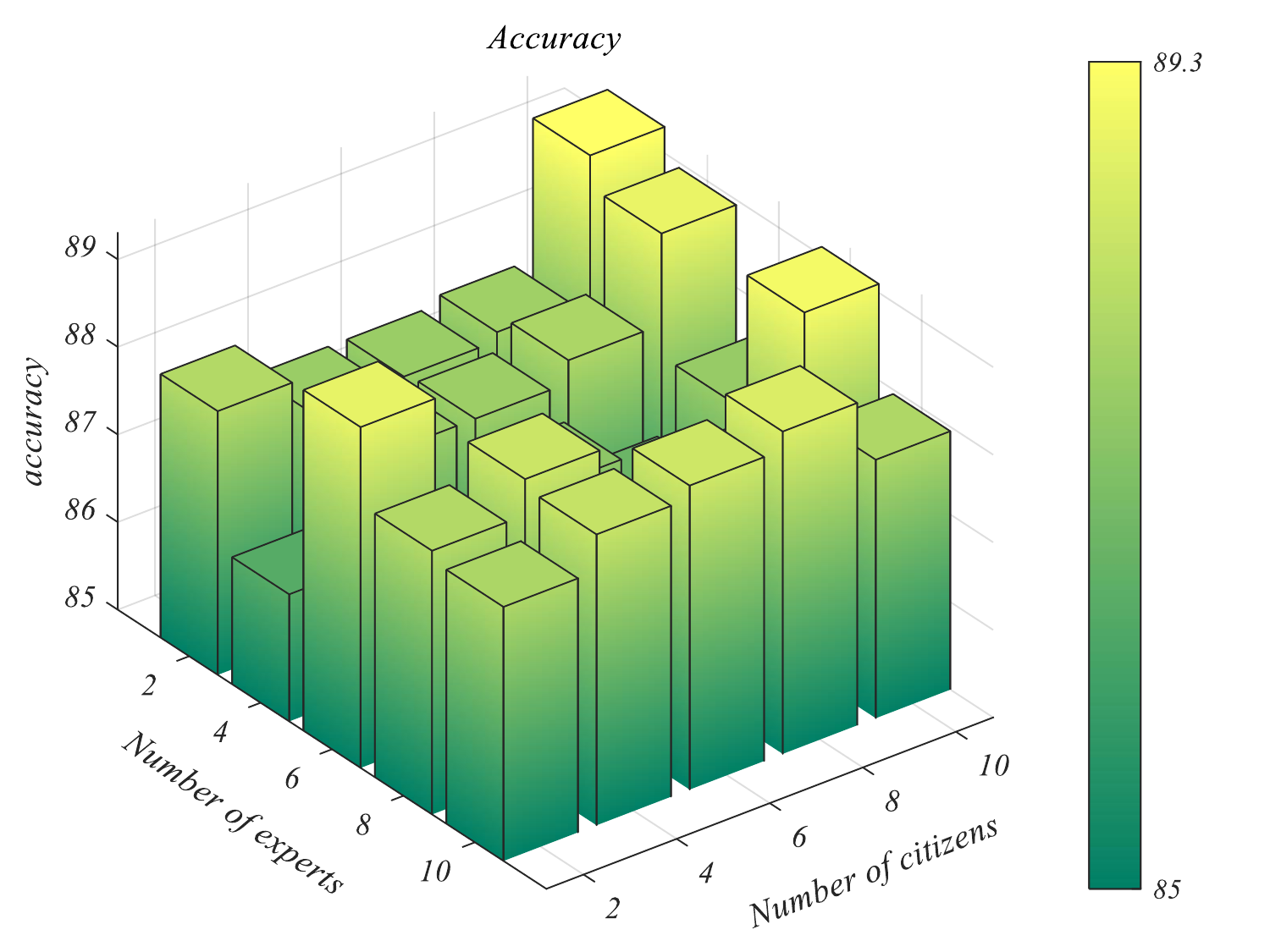}
    \caption{Parameter Sensitivity}
    \label{fig:Parameter Sensitivity}
\end{figure}
Figure \ref{fig:Parameter Sensitivity} offers a detailed insight into the sensitivity of the model's performance concerning variations in the number of experts and citizens, focusing on Accuracy as the evaluative metric. Analyzing the data reveals nuanced patterns and trends.
Starting with the number of experts, the data demonstrates an optimal Accuracy of 89.3\% when there are 10 experts. However, the Accuracy exhibits sensitivity to changes in the number of experts, with values fluctuating between 86.81\% and 88.89\% when there are 6 experts. This observation suggests that while a larger pool of experts generally contributes positively to precision, the choice of the specific number is critical for fine-tuning the model. Shifting focus to the number of citizens, a positive correlation emerges between the Accuracy and an increase in the number of citizens. For instance, with 10 experts, Accuracy shows an upward trajectory from 87.95\% (2 citizens) to 88.68\% (10 citizens). This trend underscores the valuable impact of citizen input on enhancing precision, emphasizing the significance of diverse perspectives in decision-making. Comparing different numbers of citizens for a fixed number of experts, such as 10, further highlights the importance of citizen input. Accuracy consistently improves as the number of citizens increases, showcasing the incremental value of incorporating diverse viewpoints.

\section{Conclusion}
We innovatively propose a comprehensive debunking process. Our discriminatory component demonstrates outstanding performance across rumors in multiple domains. Moreover, considering the real-time nature of rumors and the cost of fine-tuning LLM, we introduce RAG, which can address both the updating of rumor knowledge bases and eliminate the need for fine-tuning LLM while generating high-quality debunking content. Certainly, our research also has certain limitations. Currently, many rumors themselves are not problematic, but rather, the majority of manipulation occurs within accompanying images (deepfakes). Therefore, a potential direction for future exploration is in the multimodal realm, where systems can simultaneously identify inconsistencies within images and generate explanatory content accordingly.

\bibliographystyle{ACM-Reference-Format}
\bibliography{reference}


\begin{thebibliography}{38}


\ifx \showCODEN    \undefined \def \showCODEN     #1{\unskip}     \fi
\ifx \showDOI      \undefined \def \showDOI       #1{#1}\fi
\ifx \showISBNx    \undefined \def \showISBNx     #1{\unskip}     \fi
\ifx \showISBNxiii \undefined \def \showISBNxiii  #1{\unskip}     \fi
\ifx \showISSN     \undefined \def \showISSN      #1{\unskip}     \fi
\ifx \showLCCN     \undefined \def \showLCCN      #1{\unskip}     \fi
\ifx \shownote     \undefined \def \shownote      #1{#1}          \fi
\ifx \showarticletitle \undefined \def \showarticletitle #1{#1}   \fi
\ifx \showURL      \undefined \def \showURL       {\relax}        \fi
\providecommand\bibfield[2]{#2}
\providecommand\bibinfo[2]{#2}
\providecommand\natexlab[1]{#1}
\providecommand\showeprint[2][]{arXiv:#2}

\bibitem[Achiam et~al\mbox{.}(2023)]%
        {achiam2023gpt}
\bibfield{author}{\bibinfo{person}{Josh Achiam}, \bibinfo{person}{Steven Adler}, \bibinfo{person}{Sandhini Agarwal}, \bibinfo{person}{Lama Ahmad}, \bibinfo{person}{Ilge Akkaya}, \bibinfo{person}{Florencia~Leoni Aleman}, \bibinfo{person}{Diogo Almeida}, \bibinfo{person}{Janko Altenschmidt}, \bibinfo{person}{Sam Altman}, \bibinfo{person}{Shyamal Anadkat}, {et~al\mbox{.}}} \bibinfo{year}{2023}\natexlab{}.
\newblock \showarticletitle{Gpt-4 technical report}.
\newblock \bibinfo{journal}{\emph{arXiv preprint arXiv:2303.08774}} (\bibinfo{year}{2023}).
\newblock


\bibitem[Bi et~al\mbox{.}(2022)]%
        {bi2022microblog}
\bibfield{author}{\bibinfo{person}{Bei Bi}, \bibinfo{person}{Yaojun Wang}, \bibinfo{person}{Haicang Zhang}, {and} \bibinfo{person}{Yang Gao}.} \bibinfo{year}{2022}\natexlab{}.
\newblock \showarticletitle{Microblog-HAN: A micro-blog rumor detection model based on heterogeneous graph attention network}.
\newblock \bibinfo{journal}{\emph{Plos one}} \bibinfo{volume}{17}, \bibinfo{number}{4} (\bibinfo{year}{2022}), \bibinfo{pages}{e0266598}.
\newblock


\bibitem[Chen and Shu(2023)]%
        {chen2023combating}
\bibfield{author}{\bibinfo{person}{Canyu Chen} {and} \bibinfo{person}{Kai Shu}.} \bibinfo{year}{2023}\natexlab{}.
\newblock \showarticletitle{Combating Misinformation in the Age of LLMs: Opportunities and Challenges}.
\newblock \bibinfo{journal}{\emph{arXiv preprint arXiv: 2311.05656}} (\bibinfo{year}{2023}).
\newblock


\bibitem[Chen and Shu(2024)]%
        {anonymous2024can}
\bibfield{author}{\bibinfo{person}{Canyu Chen} {and} \bibinfo{person}{Kai Shu}.} \bibinfo{year}{2024}\natexlab{}.
\newblock \showarticletitle{Can {LLM}-Generated Misinformation Be Detected?}. In \bibinfo{booktitle}{\emph{The Twelfth International Conference on Learning Representations}}.
\newblock


\bibitem[Chen et~al\mbox{.}(2023)]%
        {chen2023can}
\bibfield{author}{\bibinfo{person}{Mengyang Chen}, \bibinfo{person}{Lingwei Wei}, \bibinfo{person}{Han Cao}, \bibinfo{person}{Wei Zhou}, {and} \bibinfo{person}{Songlin Hu}.} \bibinfo{year}{2023}\natexlab{}.
\newblock \showarticletitle{Can Large Language Models Understand Content and Propagation for Misinformation Detection: An Empirical Study}.
\newblock \bibinfo{journal}{\emph{arXiv preprint arXiv:2311.12699}} (\bibinfo{year}{2023}).
\newblock


\bibitem[Cui et~al\mbox{.}(2023a)]%
        {cui2023intra}
\bibfield{author}{\bibinfo{person}{Benkuan Cui}, \bibinfo{person}{Kun Ma}, \bibinfo{person}{Leping Li}, \bibinfo{person}{Weijuan Zhang}, \bibinfo{person}{Ke Ji}, \bibinfo{person}{Zhenxiang Chen}, {and} \bibinfo{person}{Ajith Abraham}.} \bibinfo{year}{2023}\natexlab{a}.
\newblock \showarticletitle{Intra-graph and Inter-graph joint information propagation network with third-order text graph tensor for fake news detection}.
\newblock \bibinfo{journal}{\emph{Applied Intelligence}} (\bibinfo{year}{2023}), \bibinfo{pages}{1--18}.
\newblock


\bibitem[Cui et~al\mbox{.}(2020)]%
        {cui2020deterrent}
\bibfield{author}{\bibinfo{person}{Limeng Cui}, \bibinfo{person}{Haeseung Seo}, \bibinfo{person}{Maryam Tabar}, \bibinfo{person}{Fenglong Ma}, \bibinfo{person}{Suhang Wang}, {and} \bibinfo{person}{Dongwon Lee}.} \bibinfo{year}{2020}\natexlab{}.
\newblock \showarticletitle{Deterrent: Knowledge guided graph attention network for detecting healthcare misinformation}. In \bibinfo{booktitle}{\emph{Proceedings of the 26th ACM SIGKDD international conference on knowledge discovery \& data mining}}. \bibinfo{pages}{492--502}.
\newblock


\bibitem[Cui et~al\mbox{.}(2023b)]%
        {cui2023efficient}
\bibfield{author}{\bibinfo{person}{Yiming Cui}, \bibinfo{person}{Ziqing Yang}, {and} \bibinfo{person}{Xin Yao}.} \bibinfo{year}{2023}\natexlab{b}.
\newblock \showarticletitle{Efficient and effective text encoding for chinese llama and alpaca}.
\newblock \bibinfo{journal}{\emph{arXiv preprint arXiv:2304.08177}} (\bibinfo{year}{2023}).
\newblock


\bibitem[Devlin et~al\mbox{.}(2018)]%
        {devlin2018bert}
\bibfield{author}{\bibinfo{person}{Jacob Devlin}, \bibinfo{person}{Ming-Wei Chang}, \bibinfo{person}{Kenton Lee}, {and} \bibinfo{person}{Kristina Toutanova}.} \bibinfo{year}{2018}\natexlab{}.
\newblock \showarticletitle{Bert: Pre-training of deep bidirectional transformers for language understanding}.
\newblock \bibinfo{journal}{\emph{arXiv preprint arXiv:1810.04805}} (\bibinfo{year}{2018}).
\newblock


\bibitem[Didem et~al\mbox{.}(2021)]%
        {2021Fake}
\bibfield{author}{\bibinfo{person}{Pehlivanoglu Didem}, \bibinfo{person}{Lin Tian}, \bibinfo{person}{Chi Kevin}, \bibinfo{person}{Perez Eliany}, \bibinfo{person}{Polk Rebecca}, \bibinfo{person}{Cahill Barian}, \bibinfo{person}{Lighthall Nichole}, {and} \bibinfo{person}{Ebner Natalie}.} \bibinfo{year}{2021}\natexlab{}.
\newblock \showarticletitle{Fake News Detection in Aging During the Era of Infodemic}.
\newblock \bibinfo{journal}{\emph{Innovation in Aging}} (\bibinfo{year}{2021}).
\newblock


\bibitem[Dou et~al\mbox{.}(2021)]%
        {dou2021user}
\bibfield{author}{\bibinfo{person}{Yingtong Dou}, \bibinfo{person}{Kai Shu}, \bibinfo{person}{Congying Xia}, \bibinfo{person}{Philip~S Yu}, {and} \bibinfo{person}{Lichao Sun}.} \bibinfo{year}{2021}\natexlab{}.
\newblock \showarticletitle{User preference-aware fake news detection}. In \bibinfo{booktitle}{\emph{Proceedings of the 44th International ACM SIGIR Conference on Research and Development in Information Retrieval}}. \bibinfo{pages}{2051--2055}.
\newblock


\bibitem[He et~al\mbox{.}(2023)]%
        {he2023efficient}
\bibfield{author}{\bibinfo{person}{Liyang He}, \bibinfo{person}{Zhenya Huang}, \bibinfo{person}{Enhong Chen}, \bibinfo{person}{Qi Liu}, \bibinfo{person}{Shiwei Tong}, \bibinfo{person}{Hao Wang}, \bibinfo{person}{Defu Lian}, {and} \bibinfo{person}{Shijin Wang}.} \bibinfo{year}{2023}\natexlab{}.
\newblock \showarticletitle{An Efficient and Robust Semantic Hashing Framework for Similar Text Search}.
\newblock \bibinfo{journal}{\emph{ACM Transactions on Information Systems}} \bibinfo{volume}{41}, \bibinfo{number}{4} (\bibinfo{year}{2023}), \bibinfo{pages}{1--31}.
\newblock


\bibitem[He et~al\mbox{.}(2020)]%
        {he2020deberta}
\bibfield{author}{\bibinfo{person}{Pengcheng He}, \bibinfo{person}{Xiaodong Liu}, \bibinfo{person}{Jianfeng Gao}, {and} \bibinfo{person}{Weizhu Chen}.} \bibinfo{year}{2020}\natexlab{}.
\newblock \showarticletitle{Deberta: Decoding-enhanced bert with disentangled attention}.
\newblock \bibinfo{journal}{\emph{arXiv preprint arXiv:2006.03654}} (\bibinfo{year}{2020}).
\newblock


\bibitem[Hu et~al\mbox{.}(2022)]%
        {hu-etal-2022-chef}
\bibfield{author}{\bibinfo{person}{Xuming Hu}, \bibinfo{person}{Zhijiang Guo}, \bibinfo{person}{GuanYu Wu}, \bibinfo{person}{Aiwei Liu}, \bibinfo{person}{Lijie Wen}, {and} \bibinfo{person}{Philip Yu}.} \bibinfo{year}{2022}\natexlab{}.
\newblock \showarticletitle{{CHEF}: A Pilot {C}hinese Dataset for Evidence-Based Fact-Checking}. In \bibinfo{booktitle}{\emph{Proceedings of the 2022 Conference of the North American Chapter of the Association for Computational Linguistics: Human Language Technologies}}, \bibfield{editor}{\bibinfo{person}{Marine Carpuat}, \bibinfo{person}{Marie-Catherine de~Marneffe}, {and} \bibinfo{person}{Ivan~Vladimir Meza~Ruiz}} (Eds.). \bibinfo{publisher}{Association for Computational Linguistics}, \bibinfo{address}{Seattle, United States}, \bibinfo{pages}{3362--3376}.
\newblock


\bibitem[Ji et~al\mbox{.}(2023)]%
        {ji2023survey}
\bibfield{author}{\bibinfo{person}{Ziwei Ji}, \bibinfo{person}{Nayeon Lee}, \bibinfo{person}{Rita Frieske}, \bibinfo{person}{Tiezheng Yu}, \bibinfo{person}{Dan Su}, \bibinfo{person}{Yan Xu}, \bibinfo{person}{Etsuko Ishii}, \bibinfo{person}{Ye~Jin Bang}, \bibinfo{person}{Andrea Madotto}, {and} \bibinfo{person}{Pascale Fung}.} \bibinfo{year}{2023}\natexlab{}.
\newblock \showarticletitle{Survey of hallucination in natural language generation}.
\newblock \bibinfo{journal}{\emph{Comput. Surveys}} \bibinfo{volume}{55}, \bibinfo{number}{12} (\bibinfo{year}{2023}), \bibinfo{pages}{1--38}.
\newblock


\bibitem[Jiang et~al\mbox{.}(2023)]%
        {jiang2023structgpt}
\bibfield{author}{\bibinfo{person}{Jinhao Jiang}, \bibinfo{person}{Kun Zhou}, \bibinfo{person}{Zican Dong}, \bibinfo{person}{Keming Ye}, \bibinfo{person}{Wayne~Xin Zhao}, {and} \bibinfo{person}{Ji-Rong Wen}.} \bibinfo{year}{2023}\natexlab{}.
\newblock \showarticletitle{Structgpt: A general framework for large language model to reason over structured data}.
\newblock \bibinfo{journal}{\emph{arXiv preprint arXiv:2305.09645}} (\bibinfo{year}{2023}).
\newblock


\bibitem[Lan et~al\mbox{.}(2019)]%
        {lan2019albert}
\bibfield{author}{\bibinfo{person}{Zhenzhong Lan}, \bibinfo{person}{Mingda Chen}, \bibinfo{person}{Sebastian Goodman}, \bibinfo{person}{Kevin Gimpel}, \bibinfo{person}{Piyush Sharma}, {and} \bibinfo{person}{Radu Soricut}.} \bibinfo{year}{2019}\natexlab{}.
\newblock \showarticletitle{Albert: A lite bert for self-supervised learning of language representations}.
\newblock \bibinfo{journal}{\emph{arXiv preprint arXiv:1909.11942}} (\bibinfo{year}{2019}).
\newblock


\bibitem[Lewis et~al\mbox{.}(2019)]%
        {lewis2019bart}
\bibfield{author}{\bibinfo{person}{Mike Lewis}, \bibinfo{person}{Yinhan Liu}, \bibinfo{person}{Naman Goyal}, \bibinfo{person}{Marjan Ghazvininejad}, \bibinfo{person}{Abdelrahman Mohamed}, \bibinfo{person}{Omer Levy}, \bibinfo{person}{Ves Stoyanov}, {and} \bibinfo{person}{Luke Zettlemoyer}.} \bibinfo{year}{2019}\natexlab{}.
\newblock \showarticletitle{Bart: Denoising sequence-to-sequence pre-training for natural language generation, translation, and comprehension}.
\newblock \bibinfo{journal}{\emph{arXiv preprint arXiv:1910.13461}} (\bibinfo{year}{2019}).
\newblock


\bibitem[Lewis et~al\mbox{.}(2020)]%
        {lewis2020retrieval}
\bibfield{author}{\bibinfo{person}{Patrick Lewis}, \bibinfo{person}{Ethan Perez}, \bibinfo{person}{Aleksandra Piktus}, \bibinfo{person}{Fabio Petroni}, \bibinfo{person}{Vladimir Karpukhin}, \bibinfo{person}{Naman Goyal}, \bibinfo{person}{Heinrich K{\"u}ttler}, \bibinfo{person}{Mike Lewis}, \bibinfo{person}{Wen-tau Yih}, \bibinfo{person}{Tim Rockt{\"a}schel}, {et~al\mbox{.}}} \bibinfo{year}{2020}\natexlab{}.
\newblock \showarticletitle{Retrieval-augmented generation for knowledge-intensive nlp tasks}.
\newblock \bibinfo{journal}{\emph{Advances in Neural Information Processing Systems}}  \bibinfo{volume}{33} (\bibinfo{year}{2020}), \bibinfo{pages}{9459--9474}.
\newblock


\bibitem[Lin et~al\mbox{.}(2023)]%
        {lin2023zero}
\bibfield{author}{\bibinfo{person}{Hongzhan Lin}, \bibinfo{person}{Pengyao Yi}, \bibinfo{person}{Jing Ma}, \bibinfo{person}{Haiyun Jiang}, \bibinfo{person}{Ziyang Luo}, \bibinfo{person}{Shuming Shi}, {and} \bibinfo{person}{Ruifang Liu}.} \bibinfo{year}{2023}\natexlab{}.
\newblock \showarticletitle{Zero-shot rumor detection with propagation structure via prompt learning}. In \bibinfo{booktitle}{\emph{Proceedings of the AAAI Conference on Artificial Intelligence}}, Vol.~\bibinfo{volume}{37}. \bibinfo{pages}{5213--5221}.
\newblock


\bibitem[Liu et~al\mbox{.}(2019)]%
        {liu2019roberta}
\bibfield{author}{\bibinfo{person}{Yinhan Liu}, \bibinfo{person}{Myle Ott}, \bibinfo{person}{Naman Goyal}, \bibinfo{person}{Jingfei Du}, \bibinfo{person}{Mandar Joshi}, \bibinfo{person}{Danqi Chen}, \bibinfo{person}{Omer Levy}, \bibinfo{person}{Mike Lewis}, \bibinfo{person}{Luke Zettlemoyer}, {and} \bibinfo{person}{Veselin Stoyanov}.} \bibinfo{year}{2019}\natexlab{}.
\newblock \showarticletitle{Roberta: A robustly optimized bert pretraining approach}.
\newblock \bibinfo{journal}{\emph{arXiv preprint arXiv:1907.11692}} (\bibinfo{year}{2019}).
\newblock


\bibitem[Nguyen et~al\mbox{.}(2020)]%
        {nguyen2020fang}
\bibfield{author}{\bibinfo{person}{Van-Hoang Nguyen}, \bibinfo{person}{Kazunari Sugiyama}, \bibinfo{person}{Preslav Nakov}, {and} \bibinfo{person}{Min-Yen Kan}.} \bibinfo{year}{2020}\natexlab{}.
\newblock \showarticletitle{Fang: Leveraging social context for fake news detection using graph representation}. In \bibinfo{booktitle}{\emph{Proceedings of the 29th ACM international conference on information \& knowledge management}}. \bibinfo{pages}{1165--1174}.
\newblock


\bibitem[Pavlyshenko(2023)]%
        {pavlyshenko2023analysis}
\bibfield{author}{\bibinfo{person}{Bohdan~M Pavlyshenko}.} \bibinfo{year}{2023}\natexlab{}.
\newblock \showarticletitle{Analysis of disinformation and fake news detection using fine-tuned large language model}.
\newblock \bibinfo{journal}{\emph{arXiv preprint arXiv:2309.04704}} (\bibinfo{year}{2023}).
\newblock


\bibitem[Pham et~al\mbox{.}(2022)]%
        {pham2022bot2vec}
\bibfield{author}{\bibinfo{person}{Phu Pham}, \bibinfo{person}{Loan~TT Nguyen}, \bibinfo{person}{Bay Vo}, {and} \bibinfo{person}{Unil Yun}.} \bibinfo{year}{2022}\natexlab{}.
\newblock \showarticletitle{Bot2Vec: A general approach of intra-community oriented representation learning for bot detection in different types of social networks}.
\newblock \bibinfo{journal}{\emph{Information Systems}}  \bibinfo{volume}{103} (\bibinfo{year}{2022}), \bibinfo{pages}{101771}.
\newblock


\bibitem[Singhal et~al\mbox{.}(2023)]%
        {singhal2023large}
\bibfield{author}{\bibinfo{person}{Karan Singhal}, \bibinfo{person}{Shekoofeh Azizi}, \bibinfo{person}{Tao Tu}, \bibinfo{person}{S~Sara Mahdavi}, \bibinfo{person}{Jason Wei}, \bibinfo{person}{Hyung~Won Chung}, \bibinfo{person}{Nathan Scales}, \bibinfo{person}{Ajay Tanwani}, \bibinfo{person}{Heather Cole-Lewis}, \bibinfo{person}{Stephen Pfohl}, {et~al\mbox{.}}} \bibinfo{year}{2023}\natexlab{}.
\newblock \showarticletitle{Large language models encode clinical knowledge}.
\newblock \bibinfo{journal}{\emph{Nature}} \bibinfo{volume}{620}, \bibinfo{number}{7972} (\bibinfo{year}{2023}), \bibinfo{pages}{172--180}.
\newblock


\bibitem[Sun et~al\mbox{.}(2023)]%
        {sun2023inconsistent}
\bibfield{author}{\bibinfo{person}{Mengzhu Sun}, \bibinfo{person}{Xi Zhang}, \bibinfo{person}{Jianqiang Ma}, \bibinfo{person}{Sihong Xie}, \bibinfo{person}{Yazheng Liu}, {and} \bibinfo{person}{S~Yu Philip}.} \bibinfo{year}{2023}\natexlab{}.
\newblock \showarticletitle{Inconsistent matters: A knowledge-guided dual-consistency network for multi-modal rumor detection}.
\newblock \bibinfo{journal}{\emph{IEEE Transactions on Knowledge and Data Engineering}} (\bibinfo{year}{2023}).
\newblock


\bibitem[Tian et~al\mbox{.}(2023)]%
        {tian2023graph}
\bibfield{author}{\bibinfo{person}{Yijun Tian}, \bibinfo{person}{Huan Song}, \bibinfo{person}{Zichen Wang}, \bibinfo{person}{Haozhu Wang}, \bibinfo{person}{Ziqing Hu}, \bibinfo{person}{Fang Wang}, \bibinfo{person}{Nitesh~V Chawla}, {and} \bibinfo{person}{Panpan Xu}.} \bibinfo{year}{2023}\natexlab{}.
\newblock \showarticletitle{Graph neural prompting with large language models}.
\newblock \bibinfo{journal}{\emph{arXiv preprint arXiv:2309.15427}} (\bibinfo{year}{2023}).
\newblock


\bibitem[Touvron et~al\mbox{.}(2023)]%
        {touvron2023llama}
\bibfield{author}{\bibinfo{person}{Hugo Touvron}, \bibinfo{person}{Louis Martin}, \bibinfo{person}{Kevin Stone}, \bibinfo{person}{Peter Albert}, \bibinfo{person}{Amjad Almahairi}, \bibinfo{person}{Yasmine Babaei}, \bibinfo{person}{Nikolay Bashlykov}, \bibinfo{person}{Soumya Batra}, \bibinfo{person}{Prajjwal Bhargava}, \bibinfo{person}{Shruti Bhosale}, {et~al\mbox{.}}} \bibinfo{year}{2023}\natexlab{}.
\newblock \showarticletitle{Llama 2: Open foundation and fine-tuned chat models}.
\newblock \bibinfo{journal}{\emph{arXiv preprint arXiv:2307.09288}} (\bibinfo{year}{2023}).
\newblock


\bibitem[Vaswani et~al\mbox{.}(2017)]%
        {vaswani2017attention}
\bibfield{author}{\bibinfo{person}{Ashish Vaswani}, \bibinfo{person}{Noam Shazeer}, \bibinfo{person}{Niki Parmar}, \bibinfo{person}{Jakob Uszkoreit}, \bibinfo{person}{Llion Jones}, \bibinfo{person}{Aidan~N Gomez}, \bibinfo{person}{{\L}ukasz Kaiser}, {and} \bibinfo{person}{Illia Polosukhin}.} \bibinfo{year}{2017}\natexlab{}.
\newblock \showarticletitle{Attention is all you need}.
\newblock \bibinfo{journal}{\emph{Advances in neural information processing systems}}  \bibinfo{volume}{30} (\bibinfo{year}{2017}).
\newblock


\bibitem[Wei et~al\mbox{.}(2021)]%
        {wei2021finetuned}
\bibfield{author}{\bibinfo{person}{Jason Wei}, \bibinfo{person}{Maarten Bosma}, \bibinfo{person}{Vincent~Y Zhao}, \bibinfo{person}{Kelvin Guu}, \bibinfo{person}{Adams~Wei Yu}, \bibinfo{person}{Brian Lester}, \bibinfo{person}{Nan Du}, \bibinfo{person}{Andrew~M Dai}, {and} \bibinfo{person}{Quoc~V Le}.} \bibinfo{year}{2021}\natexlab{}.
\newblock \showarticletitle{Finetuned language models are zero-shot learners}.
\newblock \bibinfo{journal}{\emph{arXiv preprint arXiv:2109.01652}} (\bibinfo{year}{2021}).
\newblock


\bibitem[Wei et~al\mbox{.}(2022)]%
        {wei2022chain}
\bibfield{author}{\bibinfo{person}{Jason Wei}, \bibinfo{person}{Xuezhi Wang}, \bibinfo{person}{Dale Schuurmans}, \bibinfo{person}{Maarten Bosma}, \bibinfo{person}{Fei Xia}, \bibinfo{person}{Ed Chi}, \bibinfo{person}{Quoc~V Le}, \bibinfo{person}{Denny Zhou}, {et~al\mbox{.}}} \bibinfo{year}{2022}\natexlab{}.
\newblock \showarticletitle{Chain-of-thought prompting elicits reasoning in large language models}.
\newblock \bibinfo{journal}{\emph{Advances in Neural Information Processing Systems}}  \bibinfo{volume}{35} (\bibinfo{year}{2022}), \bibinfo{pages}{24824--24837}.
\newblock


\bibitem[Work({[n.\,d.]})]%
        {workrethinking}
\bibfield{author}{\bibinfo{person}{What Makes In-Context~Learning Work}.} \bibinfo{year}{[n.\,d.]}\natexlab{}.
\newblock \showarticletitle{Rethinking the Role of Demonstrations: What Makes In-Context Learning Work?}
\newblock  (\bibinfo{year}{[n.\,d.]}).
\newblock


\bibitem[Xu et~al\mbox{.}(2021)]%
        {xu2021unified}
\bibfield{author}{\bibinfo{person}{Fan Xu}, \bibinfo{person}{Victor~S Sheng}, {and} \bibinfo{person}{Mingwen Wang}.} \bibinfo{year}{2021}\natexlab{}.
\newblock \showarticletitle{A unified perspective for disinformation detection and truth discovery in social sensing: A survey}.
\newblock \bibinfo{journal}{\emph{ACM Computing Surveys (CSUR)}} \bibinfo{volume}{55}, \bibinfo{number}{1} (\bibinfo{year}{2021}), \bibinfo{pages}{1--33}.
\newblock


\bibitem[Yang et~al\mbox{.}(2023)]%
        {yang2023rumor}
\bibfield{author}{\bibinfo{person}{Peng Yang}, \bibinfo{person}{Juncheng Leng}, \bibinfo{person}{Guangzhen Zhao}, \bibinfo{person}{Wenjun Li}, {and} \bibinfo{person}{Haisheng Fang}.} \bibinfo{year}{2023}\natexlab{}.
\newblock \showarticletitle{Rumor detection driven by graph attention capsule network on dynamic propagation structures}.
\newblock \bibinfo{journal}{\emph{The Journal of Supercomputing}} \bibinfo{volume}{79}, \bibinfo{number}{5} (\bibinfo{year}{2023}), \bibinfo{pages}{5201--5222}.
\newblock


\bibitem[Yang et~al\mbox{.}(2019)]%
        {yang2019xlnet}
\bibfield{author}{\bibinfo{person}{Zhilin Yang}, \bibinfo{person}{Zihang Dai}, \bibinfo{person}{Yiming Yang}, \bibinfo{person}{Jaime Carbonell}, \bibinfo{person}{Russ~R Salakhutdinov}, {and} \bibinfo{person}{Quoc~V Le}.} \bibinfo{year}{2019}\natexlab{}.
\newblock \showarticletitle{Xlnet: Generalized autoregressive pretraining for language understanding}.
\newblock \bibinfo{journal}{\emph{Advances in neural information processing systems}}  \bibinfo{volume}{32} (\bibinfo{year}{2019}).
\newblock


\bibitem[You et~al\mbox{.}(2019)]%
        {you2019attributed}
\bibfield{author}{\bibinfo{person}{Di You}, \bibinfo{person}{Nguyen Vo}, \bibinfo{person}{Kyumin Lee}, {and} \bibinfo{person}{Qiang Liu}.} \bibinfo{year}{2019}\natexlab{}.
\newblock \showarticletitle{Attributed multi-relational attention network for fact-checking url recommendation}. In \bibinfo{booktitle}{\emph{Proceedings of the 28th ACM International Conference on Information and Knowledge Management}}. \bibinfo{pages}{1471--1480}.
\newblock


\bibitem[Zheng et~al\mbox{.}(2023)]%
        {zheng2023rumor}
\bibfield{author}{\bibinfo{person}{Peng Zheng}, \bibinfo{person}{Zhen Huang}, \bibinfo{person}{Yong Dou}, {and} \bibinfo{person}{Yeqing Yan}.} \bibinfo{year}{2023}\natexlab{}.
\newblock \showarticletitle{Rumor detection on social media through mining the social circles with high homogeneity}.
\newblock \bibinfo{journal}{\emph{Information Sciences}}  \bibinfo{volume}{642} (\bibinfo{year}{2023}), \bibinfo{pages}{119083}.
\newblock


\bibitem[Zhu et~al\mbox{.}(2022)]%
        {zhu2022generalizing}
\bibfield{author}{\bibinfo{person}{Yongchun Zhu}, \bibinfo{person}{Qiang Sheng}, \bibinfo{person}{Juan Cao}, \bibinfo{person}{Shuokai Li}, \bibinfo{person}{Danding Wang}, {and} \bibinfo{person}{Fuzhen Zhuang}.} \bibinfo{year}{2022}\natexlab{}.
\newblock \showarticletitle{Generalizing to the future: Mitigating entity bias in fake news detection}. In \bibinfo{booktitle}{\emph{Proceedings of the 45th International ACM SIGIR Conference on Research and Development in Information Retrieval}}. \bibinfo{pages}{2120--2125}.
\newblock


\end{thebibliography}
\appendix

\end{document}